\documentclass[english]{llncs}
\usepackage[T1]{fontenc}
\usepackage[latin9]{inputenc}
\usepackage{amsmath}
\usepackage{graphicx}

\makeatletter
\usepackage {url}
\usepackage [numbers]{natbib}
\renewcommand{\fnum@figure}{\bf Fig.~\thefigure .}
\usepackage{caption}
\usepackage[labelsep=space]{caption}

\@ifundefined{showcaptionsetup}{}{%
 \PassOptionsToPackage{caption=false}{subfig}}
\usepackage{subfig}
\makeatother

\usepackage{babel}
\begin{document}

\title{Image Segmentation by Size-Dependent Single Linkage Clustering of
a Watershed Basin Graph }

\author{Aleksandar Zlateski\textsuperscript{1} and H. Sebastian Seung\textsuperscript{2}}

\institute{\textsuperscript{1}Massachusetts Institute of Technology, Cambridge,
USA\\
\texttt{zlateski@mit.edu}\\
\textsuperscript{2}Princeton University, Princeton, USA\\
\texttt{sseung@princeton.edu}}
\maketitle
\begin{abstract}
We present a method for hierarchical image segmentation that defines
a disaffinity graph on the image, over-segments it into watershed
basins, defines a new graph on the basins, and then merges basins
with a modified, size-dependent version of single linkage clustering.
The quasilinear runtime of the method makes it suitable for segmenting
large images. We illustrate the method on the challenging problem
of segmenting 3D electron microscopic brain images.

\medskip{}

\textbf{Keywords: }Watershed, image segmentation, hierarchical clustering,
electron microscopy
\end{abstract}

\section{Introduction}

Light and electron microscopy can now produce terascale 3D images
within hours {[}1, 2{]}. For segmenting such large images, efficient
algorithms are important. The watershed algorithm has linear runtime
but tends to produce severe over-segmentation, which is typically
counteracted by pre- and/or post-processing. Here we update this classic
approach, providing a new algorithm for watershed on edge-weighted
graphs, and a novel post-processing method based on single linkage
clustering modified to use prior knowledge of segment sizes.

The input is assumed to be a disaffinity graph, in which a small edge
weight indicates that the image voxels connected by the edge are likely
to belong to the same segment. Our watershed transform works by finding
the basins of attraction of steepest descent dynamics, and has runtime
that is linear in the number of disaffinity graph edges. It yields
basins similar to those of watershed cuts {[}3, 4{]}, except that
plateaus are divided between basins consistently and in a more even
way. Our post-processing starts by examining the new graph on the
basins, in which the edge connecting two basins is assigned the same
weight as the minimal edge connecting the basins in the original disaffinity
graph. Then single linkage clustering yields a hierarchical segmentation
in which the lowest level consists of the watershed basins. Each level
of single linkage clustering is a flat segmentation in which some
of the basins are merged. If we only expect to use levels above some
minimum value $T_{\min}$, then it turns out to be equivalent and
more efficient to preprocess the original disaffinity graph before
watershed by setting all edge weights below $T_{\min}$ to a common
low value. In another pre-processing step we remove the edges with
disaffinity to allow for unsegmented regions.

We also show how to modify single linkage clustering by making it
depend not only on edge weights but also on cluster size. The modification
is useful when there is prior knowledge about the size of true segments,
and is shown to have an efficient implementation because size is a
property that is guaranteed to increase with each agglomerative step.
The runtime of single linkage clustering is quasilinear in the number
of edges in the watershed basin graph.

Felzenszwalb \emph{et al. }{[}5{]} and Guimaraes \emph{et al. }{[}6{]}
have proposed efficient image segmentation methods that are quasilinear
in the number of edges in the disaffinity graph. We show that our
method produces superior results to that of {[}5{]} for the segmentation
of neural images from serial electron microscopy.

\section{Watershed Transform}

Inspired by the \emph{drop of water principle }{[}3{]} we define a
steepest descent discrete dynamics on a connected edge-weighted graph
$G=(V,E)$ with non-negative weights. A water drop travels from a
vertex to another vertex using only \emph{locally minimal} edges.
An edge $\{u,v\}$ is \emph{locally minimal} with respect to $u$
if there is no edge in $E$ incident to $u$ with lower weight. Starting
from a vertex $v_{0}$ the evolution of the system can be represented
as a \emph{steepest descent walk} $\left\langle v_{0},e_{0},v_{1},e_{1},v_{2},\dots\right\rangle $
where every edge $e_{i}$ is locally minimal with respect to $v_{i}.$
A \emph{regional minimum $M$} is a connected subgraph of $G$ such
that there is a \emph{steepest descent walk }between any pair of vertices
in $M$, and every \emph{steepest descent walk }in $G$ starting from
a vertex in $M$ will stay within $M$. A vertex $v$ belongs to the
\emph{basin of attraction} of a \emph{regional minimum $M$ }if there
exists a \emph{steepest descent walk }from $v$ to any vertex in $M$.
Note that $v$ can belong to \emph{basins of attractions} of multiple
\emph{regional minima}. In our \emph{watershed transform} we partition
$V$ into \emph{basins of attraction} of \emph{the regional minima.
}Vertices belonging to more than one \emph{basin of attraction }will
be referred to as \emph{border vertices} and will be assigned to one
of the \emph{basins} as described below.

\begin{figure}
\subfloat[]{\protect\includegraphics[scale=0.5]{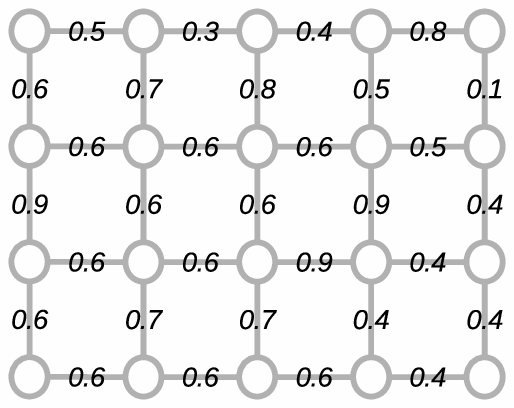}}\subfloat[]{\protect\includegraphics[scale=0.5]{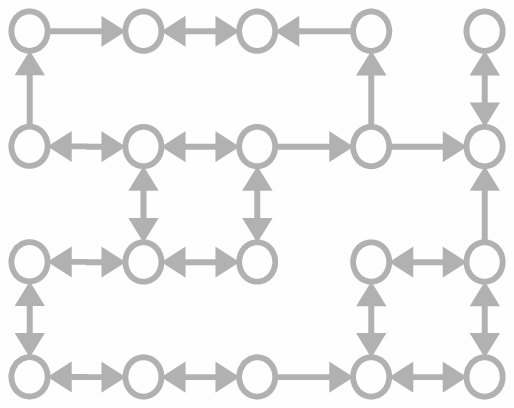}}\subfloat[]{\protect\includegraphics[scale=0.5]{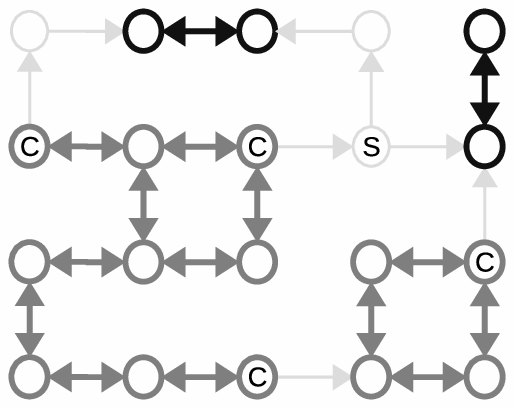}}\subfloat[]{\protect\includegraphics[scale=0.5]{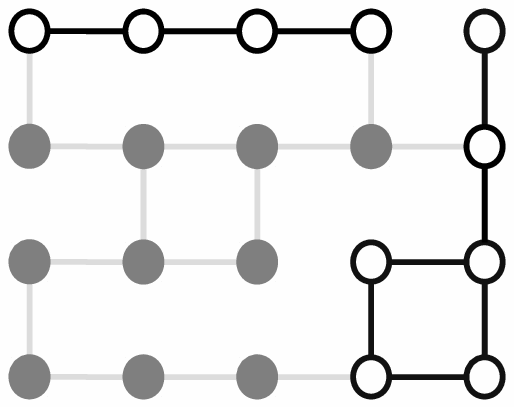}}\protect\caption{(a) A disaffinity graph; (b) derived steepest descent graph; (c) locally
minimal plateaus (black), non-minimal plateau (dark gray), saddle
vertex (S), plateau corners (C); (d) the two basins of attractions
and border vertices (dark gray)}
\end{figure}

\textbf{Steepest descent graph. }The central quantity in the watershed
algorithm is the steepest descent graph, defined as follows. Consider
an undirected weighted graph $G$ (Fig. 1(a)). Define the directed
graph $G'$ in which each undirected edge of $G$ is replaced by both
directed edges between the same vertices. The \emph{steepest descent
graph} $D$ (Fig. 1(b)) is a subgraph of $G'$ with the property that
$D$ includes every edge of $G'$ with minimal weight of all edges
outgoing from the same vertex. A directed path in $D$ is a path of
steepest descent in $G$. The steepest ascent graph can be defined
analogously using edges of maximal weight. Either steepest ascent
or descent can be used without loss of generality. For simplicity,
for a given vertex $v$ we will refer to its edges in $D$ as incoming,
outgoing, and bidirectional. A \emph{plateau} is a connected component
of the subgraph of $D$ containing only bidirectional edges. A \emph{plateau
corner} is a vertex of a \emph{plateau} that has at least one outgoing
edge. \emph{Locally minimal plateaus} contain no \emph{plateau corners,
}they\emph{ }are equivalent to the regional minima of the original
graph. \emph{Non-minimal plateaus }contain one or more \emph{plateau
corners. }A \emph{saddle vertex }has more than one outgoing edge.
In Fig. 1(c) we show \emph{locally minimal plateaus }(black), \emph{non-minimal
plateau }(dark gray), \emph{plateau corners} (C), and a\emph{ saddle
vertex }(S).

\textbf{Assigning border vertices. }In Fig. 1(d) we show the \emph{basins
of attraction} of the two \emph{regional minima.} The \emph{border}
vertices are shown in dark gray and belong to both \emph{basins of
attraction. }Watershed cuts {[}3, 4{]} assign \emph{border} vertices
with a single constraint that all the \emph{basins of attraction}
have to be connected. We introduce additional constrains. The watershed
transform has to be uniquely defined and the \emph{non-minimal plateaus
}should be divided evenly\emph{. }More specifically, we want our dynamics\emph{
}to be uniquely defined at \emph{saddle vertices, and }the vertices
of the \emph{non-minimal plateaus} to be assigned to the same \emph{basin
of attraction} as the nearest \emph{plateau corner - }a \emph{plateau
corner} reachable in fewest steps following the rules of our dynamics\emph{. }

\begin{figure}
\subfloat[]{\protect\includegraphics[scale=0.5]{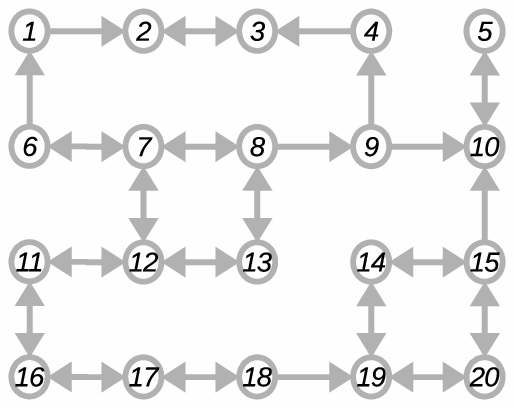}}\subfloat[]{\protect\includegraphics[scale=0.5]{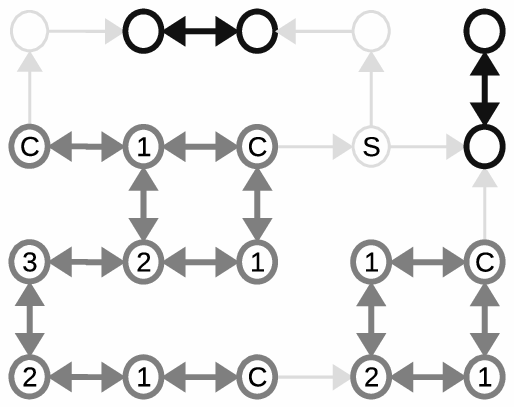}}\subfloat[]{\protect\includegraphics[scale=0.5]{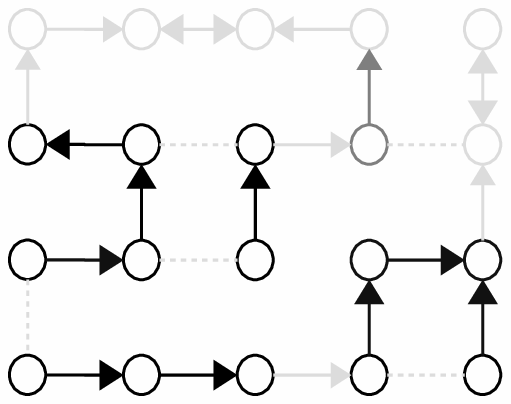}}\subfloat[]{\protect\includegraphics[scale=0.5]{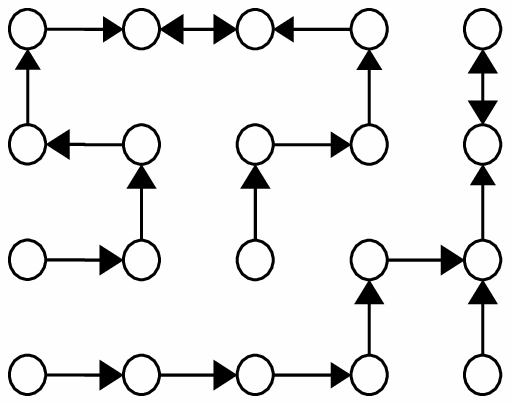}}\protect\caption{(a) Vertex indices; (b) distances to the nearest plateau corner; (c)
modifications to the steepest descent graph; (d) final watershed partition
of the graph}
\end{figure}

\textbf{Watershed transform algorithm. }We introduce an ordering function
$\alpha:V\to\{1,2,...,|V|\}$ such that $\alpha(u)\neq\alpha(v)$
if and only if $u\neq v$. We'll refer to $\alpha(u)$ as the index
of $u$ (Fig. 2(a)). In the first part of the algorithm we modify
$D$ by removing edges. For all \emph{saddle vertices} we keep only
one outgoing edge - the one pointing to a vertex with the lowest index.
In the next step we divide the \emph{non-minimal plateaus}. We initialize
a global FIFO queue $Q$, mark all the \emph{plateau corner} vertices
as visited and insert them into $Q$ in increasing order of their
index. While $Q$ is not empty we remove the vertex $v$ from the
front of the queue, we then explore all the bidirectional edges $\{v,u\}$.
If $u$ is not visited, we mark it as such, insert it to the back
of the queue and change the edge to be incoming $(v\leftarrow u)$.
Otherwise, if the vertex was already visited we just remove the edge.
The resulting steepest descent graph is shown on Fig. 2(c) - the dotted
edges are removed. Considering all the remaining edges as bidirectional,
the connected components of the modified descent graph D will be the
\emph{watershed} \emph{basins of attraction}.

The algorithm runs in linear time with respect to the number of edges
in $G$ and produces an optimal partitioning as defined in {[}3{]}.
The total number of segments in the partitioning will equal to the
total number of \emph{regional minima. }We defer the detailed algorithm
listing, the proof of correctness and running time analysis to the
supplementary material.

\textbf{Reducing over-segmentation}. Noisy values of \emph{disaffinities
}can produce severe over-segmentation (Fig. 3(c)). In order to reduce
the over-segmentation we often merge adjacent segments with the \emph{saliency
}below some given threshold $T_{\min}$ {[}7{]}. The \emph{saliency
}of two adjacent segments is defined as the value of the minimal \emph{disaffinity}
between the vertices of the two segments. That means that we are confident
that \emph{disaffinities }below $T_{\min}$ connect vertices of the
same segment. An equivalent segmentation can be obtained by replacing
the weights of all edges in $G$ with the weight smaller than $T_{\min}$
to a common low value (e.g. $0$) before applying the watershed transform.
We prove this claim in the supplementary material. To show confidence
about high values of \emph{disaffinities}, and in order to prevent
undesired mergers, we introduce a threshold $T_{\max}$ by erasing
all the edges from $G$ with the weight higher than $T_{\max}$, and
essentially setting them to $\infty$. The $T_{\max}$ threshold can
produce singleton vertices in $G$. The singleton vertices are not
assigned to any \emph{basin of attraction} and are considered background,
which is often a desired result.

\section{Hierarchical Clustering of the Watershed Basin Graph}

A hierarchical clustering of an undirected weighted graph treats each
vertex as a singleton cluster and successively merges clusters connected
by an edge in the graph. A cluster is always a connected subset of
the graph's vertices. Each merge operations creates a new level of
the hierarchy - a flat segmentation where each cluster represents
a segment. In \emph{single linkage} clustering, each step merges two
clusters connected by an edge with the lowest weight in the original
graph. \emph{Single linkage }clustering is equivalent to finding the
minimum spanning tree of the graph {[}8{]}.

In this section we propose a size-dependent single linkage clustering.
The method can be applied to any edge weighted graph, however we find
it superior when used on the \emph{watershed basin graph} defined
as follows. Let $V_{W}=\{B_{1},B_{2},\dots\}$ be the set of \emph{watershed
basins} obtained by the watershed transform of a graph $G=(V,E)$.
We define the watershed basin graph of $G$ as $G_{W}=(V_{W},E_{W})$
where an edge $\{B_{i},B_{j}\}$ exists in $E_{W}$ for all neighboring
basins $B_{i}$ and $B_{j}$, and has the weight $w(\{B_{i},B_{j}\})$
equal to the \emph{saliency} of the two \emph{basins.} We will refer
to the vertices of the \emph{watershed basin graph} as \emph{basins}
and to the edge weights as \emph{saliencies.}

In our size-dependent \emph{single linkage} clustering method, in
each step we merge clusters with the lowest \emph{saliency} that don't
satisfy a given predicate. \emph{Saliency }of two clusters is defined
as the minimal \emph{saliency }of any two members:

\begin{equation}
d_{C_{1},C_{2}}=\underset{B_{i}\in C_{1},B_{j}\in C_{2},\{B_{i},B_{j}\}\in E_{W}}{\min}w(\{B_{i},B_{j}\})
\end{equation}

At the last level of the hierarchy all pairs of clusters will satisfy
the predicate.

\textbf{Size-dependent comparison predicate. }We define a predicate
$\Lambda,$ for evaluating whether two clusters should be merged.
The predicate is based on the sizes of the two clusters. Let $S(C)$
represent the size of $C$ (e.g. number of \emph{basins} in the cluster
or the sum of the \emph{basin }sizes). We first define a non-increasing
threshold function of a cluster size $\tau(s)$. The value of $\tau(s)$
represents the maximal \emph{saliency} allowed between a cluster of
size $s$ and any adjacent cluster. Our predicate is then defined
as:

\begin{equation}
\Lambda(C_{1},C_{2})=\begin{cases}
\textrm{t\textrm{rue}} & \textrm{if }d_{C_{1},C_{2}}\ge\tau\left(\min\left\{ S(C_{1}),S(C_{2})\right\} \right)\\
\textrm{false} & \textrm{otherwise\textrm{ }}
\end{cases}
\end{equation}

The intuition behind the predicate is to apply prior knowledge about
the sizes of the true segments. With the threshold function we control
the confidence required to grow a cluster of a certain size.

With a slight modification of the predicate we could allow for an
arbitrary threshold function (changing the condition to $d_{C_{1},C_{2}}\ge\min\left\{ \tau(S(C_{1})),\tau(S(C_{2}))\right\} )$.
However, restricting the function to be non-decreasing allows us to
design a more efficient\emph{ }algorithm. It is also more intuitive
to allow higher \emph{saliency} for merging small clusters and require
lower \emph{saliency} as the sizes of the clusters grow. As $\tau$
is required to be non-decreasing, we can find a non-increasing function
$\omega$ such that when (2) is satisfied $\omega(d_{C_{1},C_{2}})\le\min\left\{ S(C_{1}),S(C_{2})\right\} $
is satisfied. This allows us to either specify either $\tau$ or $\omega$
used for the predicate. For example, when $\omega$ is constant the
algorithm will tend to aggressively merge segments smaller than the
given constant.

\textbf{Algorithm 1 }In our clustering algorithm we visit all the
edges of the watershed basin graph in non-decreasing order and merge
the corresponding clusters based on the introduced predicate.
\begin{enumerate}
\item Order $E_{W}$ into $\pi(e_{1},\dots,e_{n})$, by non-decreasing edge
weight.
\item Start with basins as singleton clusters $S^{0}=\{C_{1}=\{B_{1}\},C_{2}=\{B_{2}\},\dots\}$
\item Repeat step $4$ for $k=1,\dots,n$
\item Construct $S^{k}$ from $S^{k-1}$. Let $e_{k}=\{B_{i},B_{j}\}$ be
the $k$-th edge in the ordering. Let $C_{i}^{k-1}$ and $C_{j}^{k-1}$
be components of $S^{k-1}$ containing $B_{i}$ and $B_{j}$. If $C_{i}^{k-1}\neq C_{j}^{k-1}$
and $\Lambda(C_{i}^{k-1},C_{j}^{k-1})$ is not satisfied then $S^{k}$
is created from $S^{k-1}$ by merging $C_{i}^{k-1}$ and $C_{j}^{k-1}$,
otherwise $S^{k}=S^{k-1}.$
\item Return the hierarchical segmentation $(S^{0},\dots,S^{n})$
\end{enumerate}
\textbf{Theorem 1 }\emph{The highest level of the hierarchical segmentation
produced by algorithm (1) will have the predicate $\Lambda$ satisfied
for all pairs of the clusters. The complexity of the algorithm is
$\left|E_{W}\right|\log\left|E_{W}\right|$. The algorithm can be
modified to consider only the edges of the minimum cost spanning tree
of $G_{W}$.}

We defer the proof the supplementary material.

The steps 2-5 of the algorithm have near linear complexity. Once we
have a sorted list of the edges we can re-run the algorithm for different
threshold functions more efficiently.

\begin{figure}
\subfloat[]{\protect\includegraphics[width=3cm]{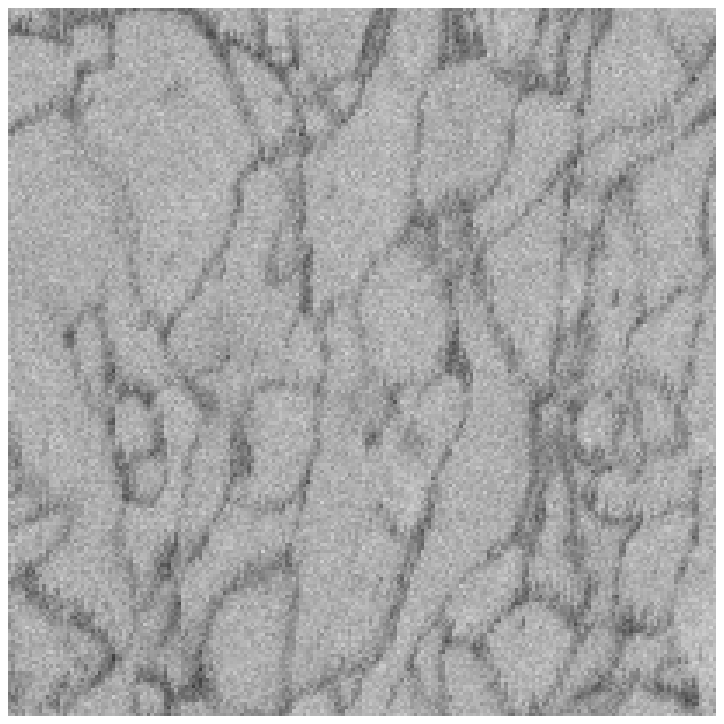}}\subfloat[]{\protect\includegraphics[width=3cm]{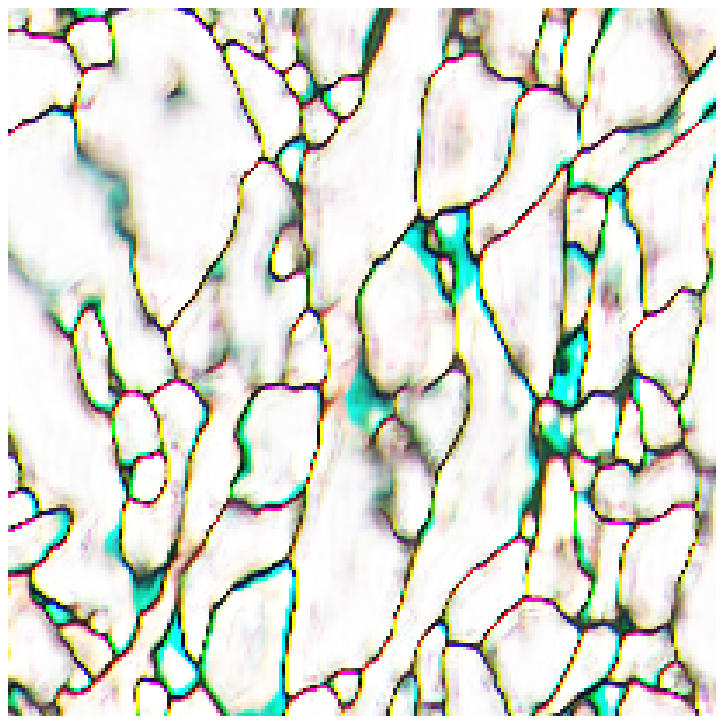}}\subfloat[]{\protect\includegraphics[width=3cm]{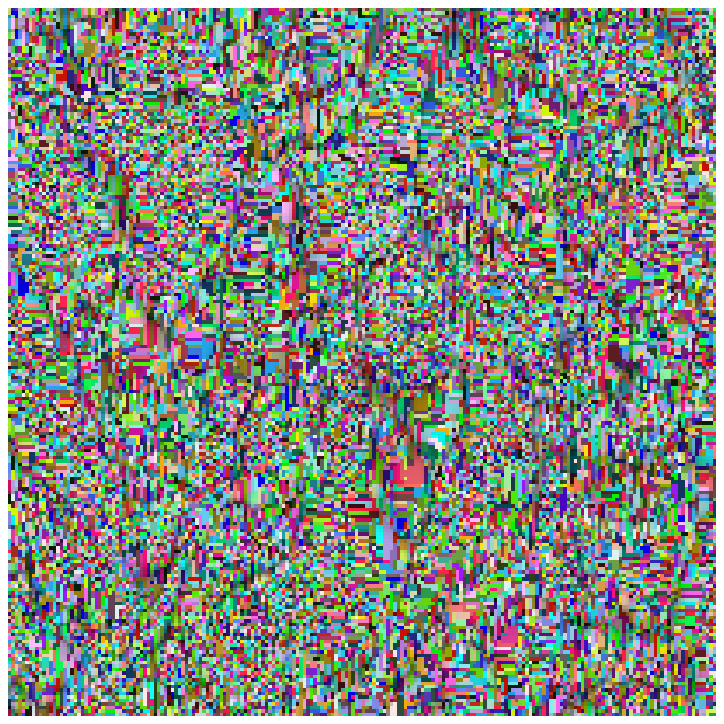}}\subfloat[]{\protect\includegraphics[width=3cm]{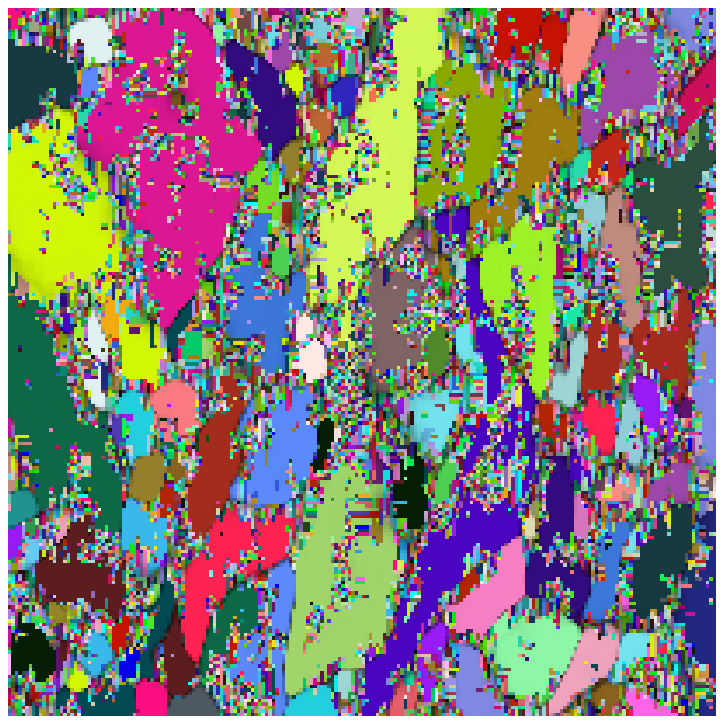}}

\subfloat[]{\protect\includegraphics[width=3cm]{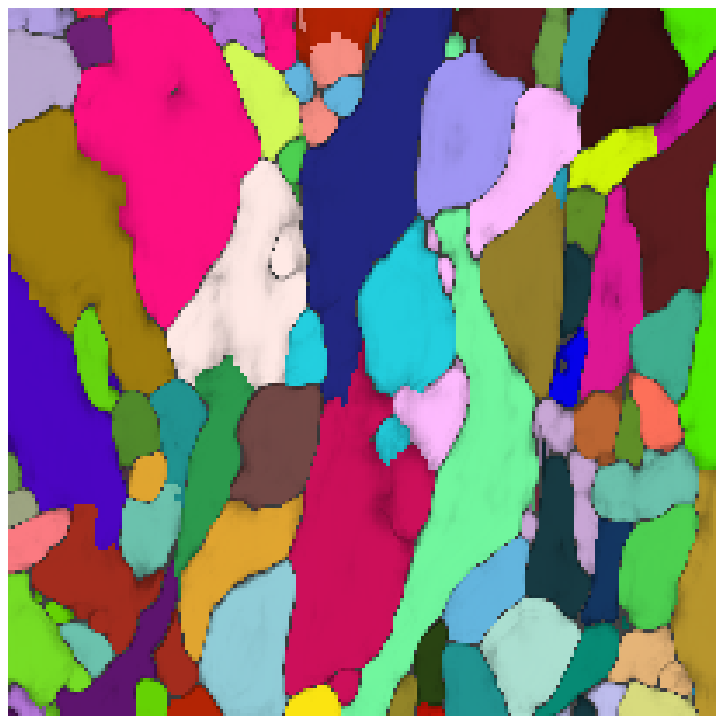}}\subfloat[]{\protect\includegraphics[width=3cm]{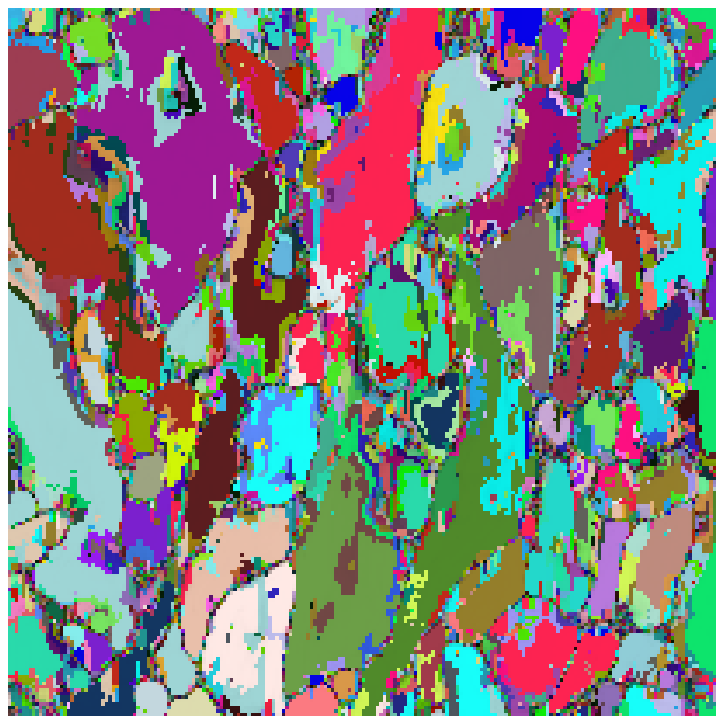}}\subfloat[]{\protect\includegraphics[width=3cm]{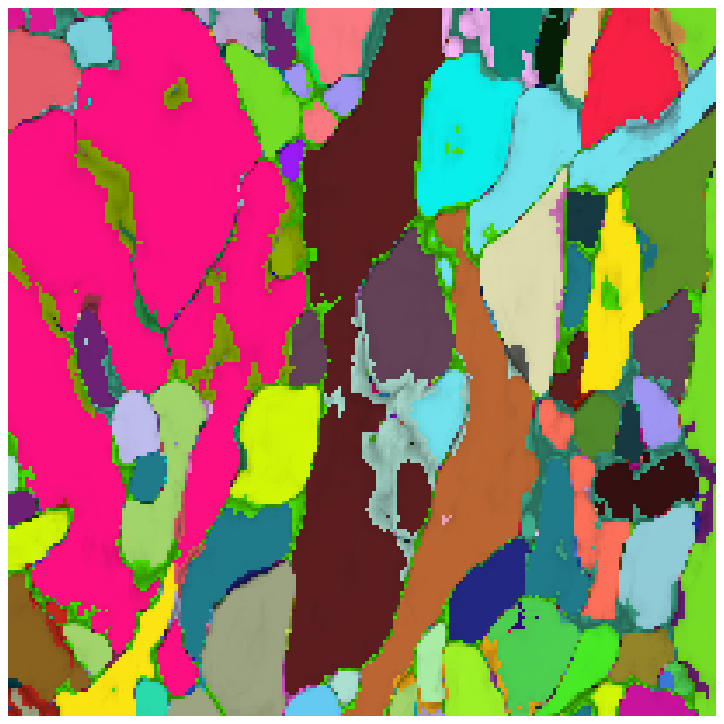}}\subfloat[]{\protect\includegraphics[width=3cm]{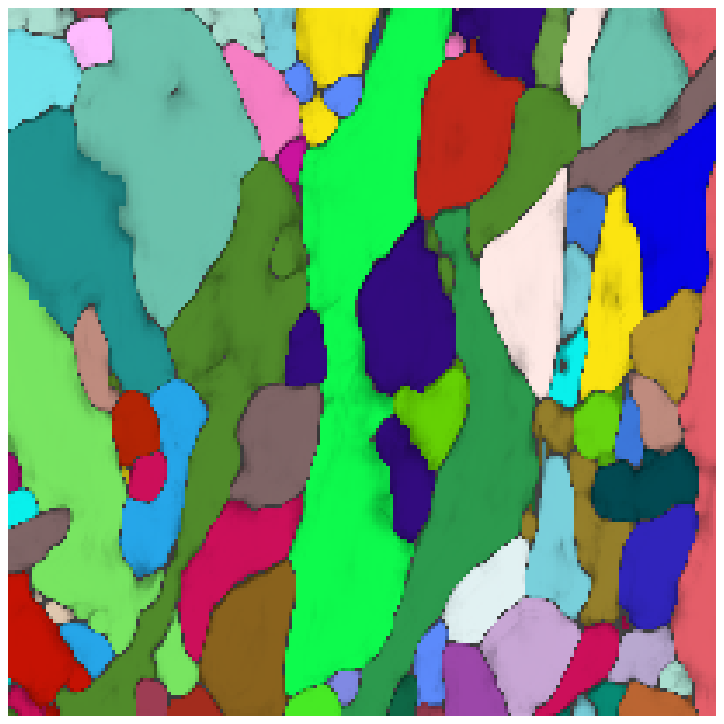}}\protect\caption{Segmentation of a $256^{3}$ EM image by our method and that of {[}5{]}
(a) slice of the raw image; (b) slice of nearest neighbor disaffinity
graph, with $xyz$ disaffinities represented with RGB; (c) watershed
transform of raw image; (d) watershed transform after preprocessing
with $T_{\min}=0.01$ and $T_{\max}=0.9$; (e) post-processing with
size-dependent single linkage clustering using $\omega(w)=3000\left(1-w\right)$;
(f, g) {[}5{]} with $k=0.5$ yields severe oversegmentation while
$k=10$ merges neurons. (h) ground truth segmentation from human expert}
\end{figure}

\section{Results}

We applied our method to 3D electron microscopic brain images {[}9{]}
(Fig. 3(a)). Disaffinity graphs were computed using convolutional
networks {[}10{]} (Fig. 3(b)). The watershed transform produced severe
oversegmentation (Fig. 3(c)), which was reduced by pre-processing
the disaffinity graph with upper and lower thresholds (Fig. 3(d)).
Size-dependent single linkage clustering further reduced oversegmentation
(Fig. 3(e)). The first function enforced all the segments to be at
least some minimal size. The second and the third functions require
the minimal size of the segment to be proportional to the affinity
(or the square of affinity).

\begin{figure}
\subfloat[]{\protect\includegraphics[width=7cm]{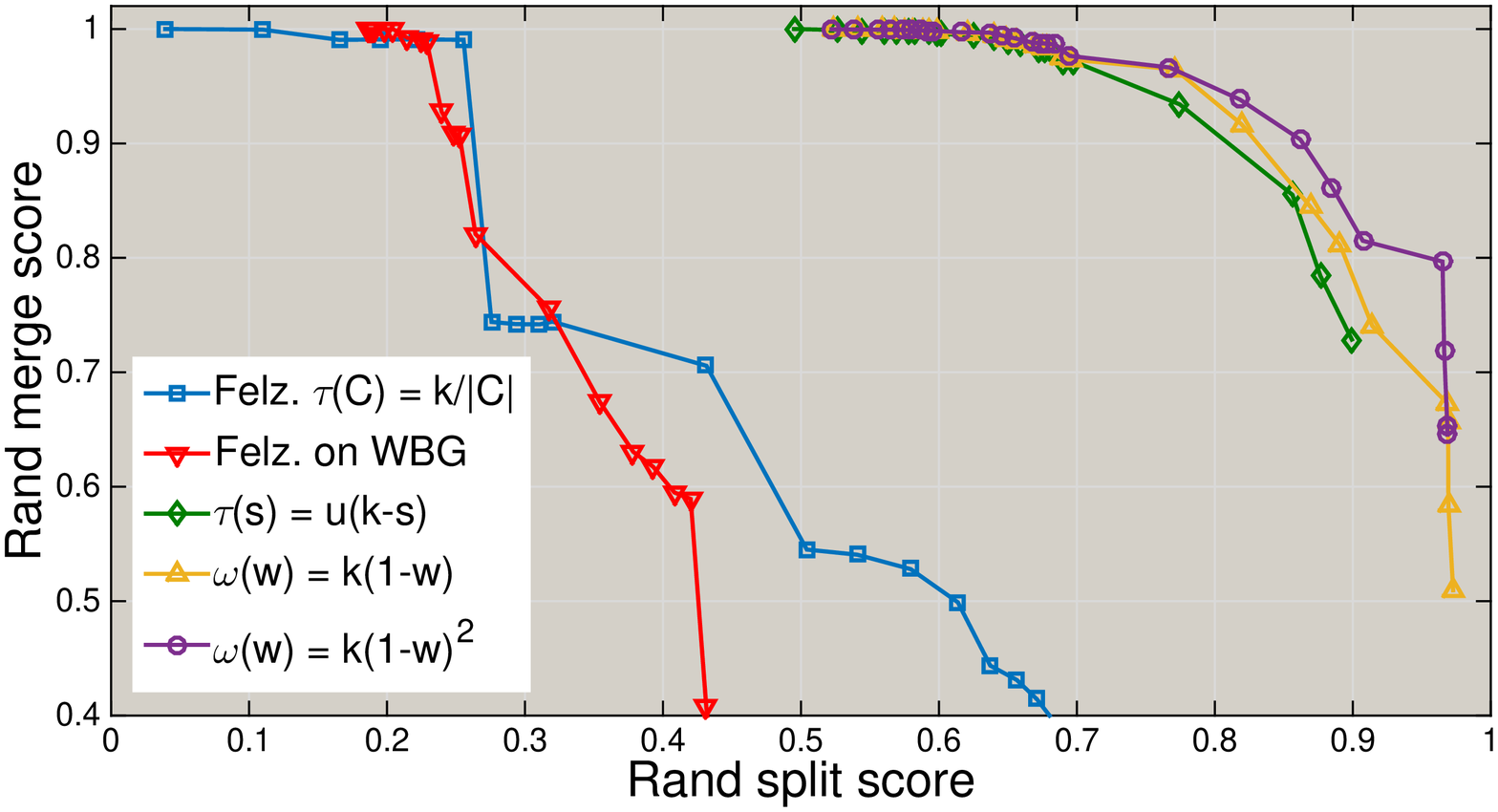}}\subfloat[]{\protect\includegraphics[width=5.5cm]{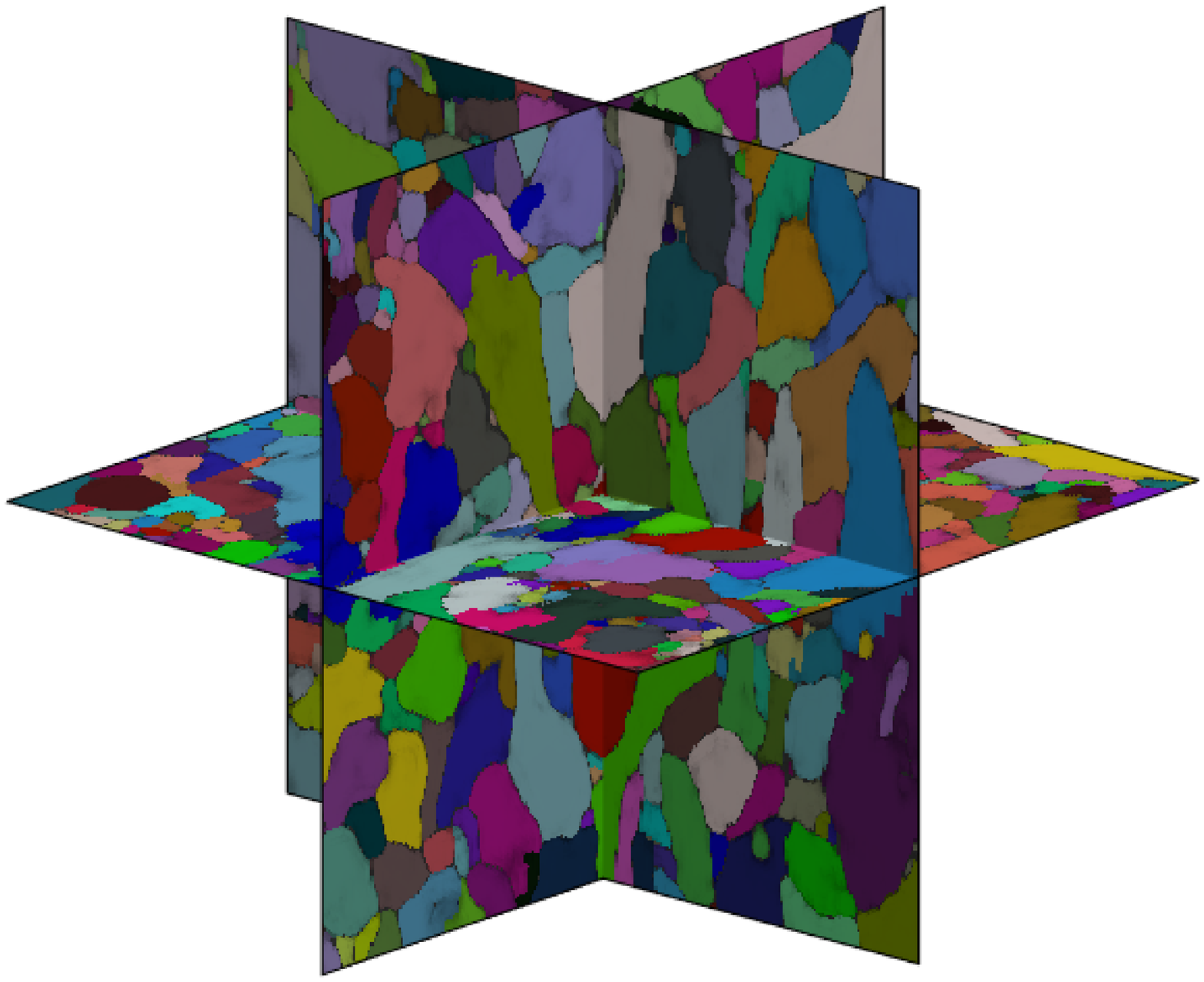}}\protect\caption{Scores of our method and that of relative to the ground truth segmentation
(a) Our method with several threshold functions, versus that of applied
to disaffinity graph and to watershed basin graph. Upper right is
better, lower left is worse; (b) Segmentation obtained by our method
with $\omega(w)=3000(1-w)$}
\end{figure}

\textbf{Measuring the quality of the segmentations. }We evaluated
the segmentations by comparing to the ground truth generated by a
human expert. Split and merge scores were computed by

\begin{equation}
V_{\textrm{split}}=\frac{{\sum_{ij}p_{ij}^{2}}}{\sum_{k}t_{k}^{2}}\textrm{ and }V_{\textrm{merge}}=\frac{{\sum_{ij}p_{ij}^{2}}}{\sum_{k}s_{k}^{2}}
\end{equation}
where $p_{ij}$ is the probability that a randomly chosen voxel belongs
to segment $i$ in the proposed segmentation and segment $j$ in the
ground truth, $s_{i}$ and $t_{j}$ are probabilities of a randomly
chosen voxel belonging to predicted segment $i$ and ground truth
segment $j$ respectively. The scores are similar to the Rand index,
a well-known metric for clustering {[}11{]}, except that they distinguish
between split and merge errors. Higher scores mean fewer errors. Scoring
was restricted to the foreground voxels in the ground truth. We tested
our method with several threshold functions, and also applied the
method of {[}5{]} to the disaffinity graph and to the \emph{watershed
basin graph}. Our method achieved superior scores (Fig. 4(a)). The
paremeter $k$ in both methods determines the trade-off between the
amount of mergers and splits. When $k$ in the method of {[}5{]} is
optimized to have approximately the same amount of mergers as our
method, large amount of splits are introduced (Fig. 3(f)) and vice
versa (Fig. 3(g)).

In conclusion, the runtime of our method makes it very suitable for
segmenting very large images. It greatly outperforms other methods
similar in runtime complexity. Our method can greatly reduce the oversegmentation
while introducing virtually no mergers.

\section*{References}
\begin{enumerate}
\item Tomer, R., Khairy, K., Amat, F., Keller, P.J.: Quantitative high-speed
imaging of entire developing embryos with simultaneous multiview lightsheet
microscopy. Nature Methods 9 755\textendash 763 (2012)
\item Marx, V.: Neurobiology: Brain mapping in high resolution. Nature 503
147\textendash 52 (2013)
\item Cousty, J., Bertrand, G., Najman, L., Couprie, M.: Watershed cuts:
Minimum spanning forests and the drop of water principle. IEEE Transactions
on Pattern Analysis and Machine Intelligence 31 1362\textendash 1374
(2009)
\item Felzenszwalb, P.F., Huttenlocher, D.P.: Efficient graph-based image
segmentation. International Journal of Computer Vision 59 167\textendash 181
(2004)
\item Guimarães, S.J.F., Cousty, J., Kenmochi, Y., Najman, L.: A Hierarchical
Image Segmentation Algorithm Based on an Observation Scale. 116\textendash{}
125 (2012)
\item Cousty, J., Bertrand, G., Najman, L., Couprie, M.: Watershed cuts:
Thinnings, shortest path forests, and topological watersheds. IEEE
Transactions on Pattern Analysis and Machine Intelligence 32 925\textendash 939
(2010)
\item Najman, L., Schmitt, M.: Geodesic Saliency of Watershed Contours and
Hierarchical Segmentation. Analysis 18 1163\textendash 1173 (1996)
\item Gower, J.C., Ross, G.J.S.: Minimum Spanning Trees and Single Linkage
Cluster Analysis. Journal of the Royal Statistical Society. Series
C (Applied Statistics) 18 54\textendash 64 (1969)
\item Briggman, K.L., Helmstaedter, M., Denk, W.: Wiring specificity in
the direction-selectivity circuit of the retina. Nature 471 183\textendash 188
(2011)
\item Helmstaedter, M., Briggman, K.L., Turaga, S.C., Jain, V., Seung, H.S.,
Denk, W.: Connectomic reconstruction of the inner plexiform layer
in the mouse retina. Nature 500 168\textendash 74 (2013)
\item Rand, W.M..: Objective Criteria for the Evaluation of Clustering Methods.
Journal of the American Statistical Association 66 846\textendash 850
(1971)\end{enumerate}

\end{document}